\documentclass[sigconf,natbib=true,anonymous=false,nonacm]{acmart}

\usepackage{makecell}
\usepackage{nicefrac}
\usepackage{todonotes}
\usepackage{xspace}

\newcommand{\wrapcenter}[1]{\begin{tabular}{c} \thead{#1} \end{tabular}}
\newcommand{\wrapcentertiny}[1]{\begin{tabular}{c}\scriptsize \thead{#1} \end{tabular}}

\AtBeginDocument{%
  \providecommand\BibTeX{{%
    \normalfont B\kern-0.5em{\scshape i\kern-0.25em b}\kern-0.8em\TeX}}}

\newif\ifpreprint
\preprinttrue

\begin{document}
\def \datasetname {\texttt{MuMiN}\xspace}
\def \trawlname {\texttt{MuMiN-trawl}\xspace}
\def \pkgname {\texttt{mumin}\xspace}
\def \dataseturl {\url{https://mumin-dataset.github.io/}}

\setlength{\tabcolsep}{1pt}

\title{\datasetname: A Large-Scale Multilingual Multimodal Fact-Checked
       Misinformation Social Network Dataset}

\author{Dan S. ~Nielsen}
\email{dan.nielsen@bristol.ac.uk}
\orcid{1234-5678-9012}
\affiliation{%
  \institution{Department of Engineering Mathematics \\ University of Bristol}
  \country{UK}
}
\author{Ryan McConville}
\email{ryan.mcconville@bristol.ac.uk}
\affiliation{%
  \institution{Department of Engineering Mathematics \\ University of Bristol}
  \country{UK}
}

\renewcommand{\shortauthors}{Nielsen and McConville}

\begin{abstract}
    Misinformation is becoming increasingly prevalent on social media and in
    news articles. It has become so widespread that we require algorithmic
    assistance utilising machine learning to detect such content. Training
    these machine learning models require datasets of sufficient scale,
    diversity and quality. However, datasets in the field of automatic
    misinformation detection are predominantly monolingual, include a limited
    amount of modalities and are not of sufficient scale and quality.
    Addressing this, we develop a data collection and linking system
    (\trawlname), to build a public misinformation graph dataset
    (\datasetname), containing rich social media data (tweets, replies, users,
    images, articles, hashtags) spanning 21 million tweets belonging to 26
    thousand Twitter threads, each of which have been semantically linked to 13
    thousand fact-checked claims across dozens of topics, events and domains,
    in 41 different languages, spanning more than a decade. The dataset is made
    available as a heterogeneous graph via a Python package (\pkgname). We
    provide baseline results for two node classification tasks related to the
    veracity of a claim involving social media, and demonstrate that these are
    challenging tasks, with the highest macro-average F1-score being 62.55\%
    and 61.45\% for the two tasks, respectively. The \datasetname ecosystem is
    available at \dataseturl, including the data, documentation, tutorials and
    leaderboards.
\end{abstract}

%





\maketitle

\section{Introduction}
\label{sec:introduction}

While it may be possible to track the history of misinformation, or `fake
news', back to Octavian of the Roman Republic \citep{watson2018information}, or
Browne in the 17th century \citep{browne1646pseudodoxia}, it was the World Wide
Web and the rise of online social networks that has provided new and powerful
ways for the rapid dissemination of information, both true and false, with
false information having negative effects across many aspects of society, such
as politics and health.

A universal consensus has yet to be reached on the definition of
misinformation. One convincing definition of misinformation is that it is
`false or misleading' information, with a subcategory of misinformation being
disinformation, which is misinformation with the intention to deceive
\citep{lazer2018fakenewsscience}. In this work, we do not specifically
distinguish disinformation from misinformation.

There exist over one hundred fact checking organisations that manually verify
the veracity of claims made online, often within news articles or social media
posts.  This is a time consuming task involving a multitude of different
documents and sources in order to classify a claim. To build intelligent tools
to help with this process,  datasets that accurately represent misinformation
are required.

Online misinformation is multimodal, multilingual and multitopical. The
multimodal aspect of misinformation manifests online in the use of image and
video in addition to the commonly studied textual communication. Moreover, we
also posit that an additional modality is the social behaviour of users online,
which exhibits itself in the form of a graph, or network, of interactions and
behaviours. These interactions vary by platform, but on Twitter, they can be
considered actions such as `retweeting', `quote tweeting' or `replying' to a
tweet, or `following' a user. While research typically studies such modalities
in isolation, or occasionally, some subset, the integration of all modalities
may be necessary to accurately capture the underlying classification of
misinformation.

The multilingual dimension of misinformation can be challenging due to the
focus of existing research on misinformation within the English language, with
most existing misinformation datasets only covering the English language. See
Table~\ref{tab:datasets} for an overview of existing misinformation datasets.
Further, there exists a focus of natural language processing research towards
the English language. Indeed, in Sections~\ref{sec:claimdataprocessing} and
\ref{sec:datalinking} we observe that the transformer based models perform
better when text is translated to English when compared to a multilingual
model.

Finally, misinformation has the potential to permeate across all aspects of
society and life, and is not restricted to a single topic or domain. For
example, a significant number of tweeted articles analysed on Twitter, in the
months preceding the 2016 United States presidential election, contained fake
or extremely biased news \citep{bovet2019fakenews}. During the COVID-19
pandemic, the World Health Organization director-general stated that not only
are they fighting the COVID-19 pandemic but also an `infodemic'
\citep{lancet2020infodemic}. Naming but two examples, this alone provides
further motivation that automated misinformation detection systems must not be
trained on a single topic (e.g., COVID-19) or domain (e.g., politics), and
thus justifies the collection of datasets that capture the pervasiveness of
misinformation across many aspects of society.

Given the severity of online misinformation, there have been numerous public
datasets made available for researchers to develop and evaluate automated
misinformation detection models with. These publicly available datasets cover
topics ranging from celebrities \citep{perez2018automatic}, rumours
\citep{zubiaga2017exploiting}, politics \citep{shu2020fakenewsnet} and health
\citep{li2020mmcovid}. These datasets typically include data from a social
network, usually Twitter, along with labels assigning them to a category,
categorizing them as some equivalent of `true' or `false'. These labels often
come from `fact-checking' resources such as
PolitiFact\footnote{\url{https://www.politifact.com/}} and
Poynter\footnote{\url{https://www.poynter.org/ifcn/}}.

There are, however, a number of limitations of existing datasets. We believe
that in order to make advances on the development of automated misinformation
detection systems, datasets that capture the breadth, complexity and scale of
the problem are required. Specifically, we believe that an effective dataset
should be large scale, as misinformation is an extremely varied and wide
ranging phenomenon, with thousands of manually fact checked claims available
online from fact-checking organisations across a range of topics. To ensure
that misinformation detection models are able to generalise to new events, we
need models to be able to learn event-independent predictors of misinformation.
We believe that such predictors will not be possible from the claim texts
alone, as they are inherently event-dependent. Instead, we argue that models
(and thus datasets to train them) should utilise the context of the claim, for
example, the social network surrounding the claim, or the article in which the
claim was posted.

Given the short message length of posts on Twitter, we believe that additional
context is required in order to properly capture how claims are being discussed
on social networks. This can take the form of the media involved in the posts,
articles that users are sharing on the social network, or indeed the social
network of the user themselves (i.e. who they follow, who follows them), as
well as interactions with these posts, such as replies or retweets. Therefore,
we semantically link fact-checked claims not only to the social network posts,
but also to this additional information. Further, given that misinformation is
a global challenge, a useful dataset should not be limited to a single
language, and should contain data in as many languages as possible.

Further, most misinformation datasets consist of a single data dump which,
given the dynamic nature of the problem, means that datasets can become
outdated. Therefore, we open source our data collection and linking
infrastructure which connects claims to social networks, \trawlname, in order
to provide a platform for future research to continue to build and extend our
work.

We see the goal of an automatic misinformation detection system as a tool that
can help people identity misinformation so that they can act on it accordingly.
Considering that a lot of the misinformation today is spread on social media
networks, such a system should be able to retrieve, connect and utilise the
information in these networks to identify misinformation as accurately as
possible. This is the core rationale behind our proposed two tasks, which we
further discuss in Section~\ref{sec:baselinemodels}:
\begin{enumerate}
    \item Determine the veracity of a claim, given its social network
        context.
    \item Determine the likelihood that a social media post to be
        fact-checked is discussing a misleading claim, given its social network
        context.
\end{enumerate}

To this end, we present \datasetname , which addresses the
limitations of existing work. In summary, our main contributions are as
follows:
\begin{itemize}
    \item We release a graph dataset, \datasetname, containing rich social
        media data (tweets, replies, users, images, articles, hashtags)
        spanning 21 million tweets belonging to 26 thousand Twitter threads,
        each of which have been semantically linked to 13 thousand fact-checked
        claims across dozens of topics, events and domains, in 41 different
        languages, spanning more than a decade.
    \item We release the data collection and linking system,\newline\trawlname,
        which was used to build the \datasetname dataset.
    \item We release a Python package, \pkgname, which eases the compilation of
        the dataset as well as enabling easy export to the Deep Graph Learning
        framework \cite{wang2019dgl}.
    \item We propose two representative tasks involving claims and social
        networks. We provide baseline results considering both text-only
        models, image-only models as well as using a heterogeneous graph neural
        network.
\end{itemize}

\setlength{\tabcolsep}{2pt}
\begin{table*}[ht!]
    \caption{The statistics of the three datasets.}
    \begin{center}
        \begin{tabular}{c|cccccccc}
            \toprule
            Dataset & \#Claims & \#Threads & \#Tweets & \#Users & \#Articles &
                \#Images & \#Languages & \%Misinfo \\
            \midrule
            \datasetname-large & 12,914 & 26,048 & 21,565,018 &
                1,986,354 & 10,920 & 6,573 & 41 & 94.79\% \\
            \datasetname-medium & 5,565 & 10,832 & 12,659,371 &
                1,150,259 & 4,212 & 2,510 & 37 & 94.20\% \\
            \datasetname-small & 2,183 & 4,344 & 7,202,506 &
                639,559 & 1,497 & 1,036 & 35 & 92.71\% \\
            \bottomrule
        \end{tabular}
        \label{tab:statistics}
    \end{center}
\end{table*}

\section{Related Work}
\label{sec:relatedwork}
There is a number of publicly available datasets on the topic of
misinformation. Some datasets have a narrow focus on topics, for example on
politics \citep{wang2017liar}, COVID-19 \citep{li2020mmcovid, cui2020coaid}, or
celebrities ~\citep{perez2018automatic}, but others, such as the
\texttt{PHEME5} \citep{zubiaga2017exploiting} and \texttt{PHEME9}
\citep{kochkina2018allinone} datasets, have explicitly sought to include data
from different events, although typically with much smaller numbers of events,
claims and tweets than \datasetname.


A popular approach is to make extensive use of fact-checking websites in order
to provide ground truth labels for misinformation datasets. Popular fact
checking data sources include
PolitiFact\footnote{\url{https://www.politifact.com}}, which has been used by
numerous datasets such as \texttt{Liar} \citep{wang2017liar}, which has around
twelve thousand facts of a political nature and associated short statements.
Others have used PolitiFact instead with news articles, such as those from
\texttt{FakeNewsNet} \citep{shu2017exploiting, shu2020fakenewsnet}. A more
recent dataset from \texttt{FakeNewsNet}, \citep{shu2020fakenewsnet}, links
articles to claims, along with tweets sharing the referenced articles. In both
cases, the number labelled news articles are, again, fewer than \datasetname,
with 240 news articles labelled in \citep{shu2017exploiting} and 1,056 news
articles labelled in \citep{shu2020fakenewsnet}.

Other work has sought to extend the number of fact-checking organisations used
to construct datasets such as the \texttt{CoAID} \citep{cui2020coaid} and
\texttt{MM}-\texttt{COVID} \citep{li2020mmcovid} datasets, which contain claims
from 6 and 97 fact checking organisations, respectively. Of the 97 used by
\texttt{MM}-\texttt{COVID}, 96 of them come from the Poynter fact-checking
network of fact-checking organisations. In total they have over 4,000 claims.
\texttt{MM}-\newline
\texttt{COVID} is the first to address the monolingual problem with
existing datasets by including data in 6 different languages, albeit on a
single topic. This dataset is perhaps the most related work to ours in that it
addresses several of the problems we outline. However, our dataset,
\datasetname, is significantly larger in terms of the number of claims, tweets,
languages and indeed topics, as we do not limit our dataset to COVID-19 related
misinformation. Further, we use a more sophisticated procedure to link claims
to tweets; the \texttt{MM}-\texttt{COVID} dataset links an article to social
media posts by searching for the URL, title and first sentence of the article
on Twitter, while our dataset performs linking based on cosine similarity
between transformer based embeddings of the claims, tweets and articles.

One important aspect of this line of research to consider is around evidence
based fact checking approaches. This line of research seeks to utilise
available evidence, such as online news sources, Wikipedia, as well as social
networks, in order to classify a claim as true or false. Research by
\citet{popat20216} proposes a system to replace manual verification of claims
with such a system. Using only the text of the claim, they retrieve online
search results using the text, and use linguistic features of the resulting
articles, as well as the reliability of the sources in order to classify a
claim. To deal with the variability in labels given by fact checkers, such as
`mostly true' or `partially false', they map `mostly' to true, and ignore those
of a `partial' veracity.

Another weakness of these approaches are that they tend to be limited in the
fact checkers that they utilise for their sources. The \texttt{MultiFC} dataset
\citep{Augenstein2019MultiFC} seeks to address this by including claims from 26
fact checking sources, in English, producing a dataset of 34,918 claims. While
they do include extra context and metadata, they do not include additional
modalities (such as images), nor do they include social network data. Recent
work by \citet{hanselowski2019} released a dataset, that while containing
documents retrieved from different domains, such as blogs, social networks and
news websites, as with other work in this area, theirs has a strong focus on
text, overlooking the relevant and rich information contained in other
modalities such as images, videos and social graphs of interaction. The same
sentiment applies also to the very recent \texttt{X-FACT} dataset
\citep{gupta2021xfact}, that while multilingual (25 languages), contains only
text data.

While significant attention has been paid to the use of fact checking
organisations as a source of ground truth for claim veracity and verification,
there has also been work studying artificial claims. One such dataset is
\texttt{FEVER} \citep{thorne2018fever} which consists of 185,445 claims created
by manipulating sentences from Wikipedia, and then annotated manually into one
of three categories, supported, refuted or not enough information. Also using
Wikipedia is the \texttt{HoVer} dataset \citep{jiang2020hover}, which addresses
a weakness of the \texttt{FEVER} dataset, in that to verify claims, often a
single Wikipedia article is not enough, and often requires multiple sources, or
`hops'. In \texttt{HoVer}, claims can require evidence from up to four
Wikipedia articles. However, \texttt{HoVer} is still a monolingual dataset with
an emphasis on text data, differing significantly from \datasetname, which
considers multiple modalities across multiple languages as important
characteristics to consider for this problem.

See Table~\ref{tab:datasets} for an overview of these datasets, which
demonstrates the key differences between them.

\setlength{\tabcolsep}{5pt}
\begin{table*}[h]
    \footnotesize
    \caption{An overview of publicly available datasets for automatic
             misinformation detection, ordered by release date. Here $\dagger$
             indicates that the tweet content is not available but that the
             related users are, and parentheses indicate that it only holds for
             a subset of the dataset.}
    \begin{center}
        \begin{tabular}{p{2.5cm}|cccccccccc}
            \toprule
            \textbf{Dataset} & \textbf{\#Facts} & \textbf{\#Tweets} &
            \textbf{Verified} & \textbf{Multilingual} & \textbf{Multitopical} &
            \textbf{Articles} & \textbf{Images} & \textbf{User} &
            \textbf{Social} & \textbf{Replies} \\

            \midrule

            \texttt{MediaEval15}~\citep{boididou2015verifying} &
                413 &  
                15,821 &  
                \checkmark &  
                &  
                \checkmark &  
                &  
                \checkmark &  
                \checkmark &  
                &  
                \\  

            \texttt{MediaEval16}~\citep{boididou2016verifying} &
                542 &  
                18,049 &  
                \checkmark &  
                &  
                \checkmark &  
                &  
                \checkmark &  
                \checkmark &  
                &  
                \\  

            \texttt{Liar}~\citep{wang2017liar} &
                12,836 &  
                &  
                \checkmark &  
                &  
                &  
                &  
                &  
                \checkmark &  
                &  
                \\  

            \texttt{Weibo}~\citep{jin2017multimodal} &
                &  
                9,528 &  
                \checkmark &  
                &  
                \checkmark &  
                &  
                \checkmark &  
                \checkmark &  
                &  
                \\  

            \texttt{PHEME5}~\citep{zubiaga2017exploiting} &
                &  
                5,802 &  
                \checkmark &  
                &  
                \checkmark &  
                &  
                &  
                &  
                &  
                \checkmark \\  

            \midrule

            \texttt{FNN-BuzzFeed}~\citep{shu2017exploiting} &
                182 &  
                &  
                \checkmark &  
                &  
                &  
                \checkmark &  
                &  
                \checkmark &  
                \checkmark &  
                \\  

            \texttt{FNN-PolitiFact17}~\citep{shu2017exploiting} &
                240 &  
                &  
                \checkmark &  
                &  
                &  
                \checkmark &  
                &  
                \checkmark &  
                \checkmark &  
                \\  

            \texttt{PHEME9}~\citep{kochkina2018allinone} &
                &  
                6,425 &  
                \checkmark &  
                &  
                \checkmark &  
                &  
                &  
                &  
                &  
                \checkmark \\  

            \texttt{Celebrity}~\citep{perez2018automatic} &
                200 &  
                &  
                \checkmark &  
                &  
                &  
                \checkmark &  
                &  
                &  
                &  
                \\  

            \texttt{FakeNewsAMT}~\citep{perez2018automatic} &
                240 &  
                &  
                &  
                &  
                \checkmark &  
                \checkmark &  
                &  
                &  
                &  
                \\  

            \texttt{FEVER}~\citep{thorne2018fever} &
                185,445 &  
                &  
                &  
                &  
                \checkmark &  
                &  
                &  
                &  
                &  
                \\  

            \midrule

            \texttt{AFCSDC}~\citep{baly2018} &
                422 &  
                &  
                \checkmark &  
                &  
                \checkmark &  
                \checkmark &  
                &  
                &  
                &  
                \\  

            \texttt{UKP Snopes}~\citep{hanselowski2019} &
                6,422 &  
                &  
                \checkmark &  
                &  
                \checkmark &  
                \checkmark &  
                &  
                &  
                &  
                \\  

            \texttt{MultiFC}~\citep{Augenstein2019MultiFC} &
                34,918 &  
                &  
                \checkmark &  
                &  
                \checkmark &  
                \checkmark &  
                &  
                &  
                &  
                \\  

            \texttt{HoVer}~\citep{jiang2020hover} &
                26,000 &  
                &  
                &  
                &  
                \checkmark &  
                &  
                &  
                &  
                &  
                \\  

            \texttt{FNN-PolitiFact20}~\citep{shu2020fakenewsnet} &
                1,056 &  
                564,129 &  
                \checkmark &  
                &  
                &  
                \checkmark &  
                \checkmark &  
                \checkmark &  
                &  
                \checkmark \\  

            \texttt{FNN-GossipCop}~\citep{shu2020fakenewsnet} &
                22,140 &  
                1,396,548 &  
                \checkmark &  
                &  
                &  
                \checkmark &  
                \checkmark &  
                \checkmark &  
                &  
                \checkmark \\  

            \midrule

            \texttt{CoAID}~\citep{cui2020coaid} &
                4,251 &  
                160,667 &  
                (\checkmark) &  
                &  
                &  
                \checkmark &  
                &  
                &  
                &  
                \checkmark \\  

            \texttt{MM-COVID}~\citep{li2020mmcovid} &
                11,565 &  
                105,300 &  
                (\checkmark) &  
                \checkmark &  
                &  
                \checkmark &  
                \checkmark &  
                \checkmark &  
                \checkmark &  
                \checkmark \\  

            \texttt{UPFD-POL}~\citep{dou2021upfd} &
                314 &  
                40,740$^\dagger$ &  
                \checkmark &  
                &  
                &  
                \checkmark &  
                &  
                \checkmark &  
                &  
                \checkmark$^\dagger$ \\  

            \texttt{UPFD-GOS}~\citep{dou2021upfd} &
                5,464 &  
                308,798$^\dagger$ &  
                \checkmark &  
                &  
                &  
                \checkmark &  
                &  
                \checkmark &  
                &  
                \checkmark$^\dagger$ \\  

            \texttt{X-FACT}~\cite{gupta2021xfact} &
                31,189 &  
                &  
                \checkmark &  
                \checkmark &  
                \checkmark &  
                &  
                &  
                &  
                &  
                \\  

            \midrule

            \datasetname &
                12,914 &  
                21,565,018 &  
                \checkmark &  
                \checkmark &  
                \checkmark &  
                \checkmark &  
                \checkmark &  
                \checkmark &  
                \checkmark &  
                \checkmark \\  

            \bottomrule
        \end{tabular}

        \label{tab:datasets}
    \end{center}
\end{table*}

\section{Dataset Creation}
\label{sec:datasetcreation}

The dataset creation consists of two parts, one concerning the claims and their
fact-checked verdicts, and the second part concerning the collection of the
surrounding social context. As described in Section~\ref{sec:introduction},
this lends itself to two application tasks, the first being, given a claim,
predict its veracity given the social context. The second being, given a
Twitter post to be fact-checked and its social context, predict the veracity of
the claim made in the Twitter post. The general strategy is to collect claims
as spatiotemporally diverse as possible, and to collect as many high-quality
social features surrounding these as possible. The dataset creation was
performed using \trawlname on a single workstation with an Intel Core i9-9900K
CPU, 64GB of RAM, with two Nvidia 2080Ti GPUs, with the collection taking
several months. Baseline results were produced on the same workstation.

\subsection{Claims}
\label{sec:claims}

\subsubsection{Data Collection}
\label{sec:claimcollection}
For the collection of fact-checked claims we utilise the Google Fact Check
Tools
API\footnote{\url{https://developers.google.com/fact-check/tools/api/reference/rest}},
which is a resource that collects fact-checked claims from fact-checking
organisations around the world. This API was also used in
\citep{shiao2021ki2te} to create a dataset for automatic misinformation
detection, but our aim was to collect a much larger amount of claims that were
sufficiently diverse, both in terms of content and language.

We started by querying the API for the queries \texttt{coronavirus} and
\texttt{covid} to ensure that we got results from active fact-checking
organisations. To ensure language diversity, we used the Google Translation
API\footnote{\url{https://cloud.google.com/translate/docs/reference/rest/}} to
translate the two queries into 70 languages\ifpreprint (see the appendix for a
list of all the languages)\fi. We then queried the Fact Check API for up to
1,000 fact-checked claims for each of the resulting 132 queries.

From the collected fact-checked claims, we collected all the fact-checking
organisations involved, resulting in a list of 115 fact-checking
organisations\ifpreprint (see the appendix for a list of all the
organisations)\fi. From this list, we scraped all the fact-checked claims from
each of them, from the fact-checking organisation's inception up until present
day. This resulted in 128,070 fact-checked claims.

\subsubsection{Data Processing}
\label{sec:claimdataprocessing}
The claim data collected from the procedure in
Section~\ref{sec:claimcollection} also included various metadata, and we
extracted the following: (1) \texttt{source}: The source of the claim, which
can be both names of people as well as generic descriptions such as ``multiple
social media posts''; (2) \texttt{reviewer}: The URL of the fact-checking
website that reviewed the claim; (3) \texttt{language}: The language the claim
was uttered or written in; (4) \texttt{verdict}: The fact-checking
organisation's verdict; (5) \texttt{date}: The date the claim was uttered. If
this date was not available then the date of the review was used. If neither of
those two were available then we extracted a potential date from the URL of the
fact-checking article using the regular expression
\texttt{[0-9]\{4\}-[0-9]\{2\}-[0-9]\{2\}} \footnote{This regular expression
matches four, two and two numbers, separated by dashes. Thus,
\texttt{2020-01-30} would be matched, but \texttt{20-01-30} would not.}.
Note, from this, we release only the date, keywords from the claim, the
predicted verdict (using the verdict classification model described below)
and the reviewer involved.

The first challenge is that the \texttt{verdict} is unstructured freetext and
can be written in any language at any length. To remedy this, we trained a
`verdict classifier', a machine learning model that classifies the freetext
verdicts into three pre-specified categories: \texttt{misinformation},
\texttt{factual} and \texttt{other}. Towards this, we manually labelled 2,500
unique verdicts. Aside from the simple classifications such as labelling
``false'' and ``misleading'' as \texttt{misinformation}, and labelling ``true''
and ``correct'' as \texttt{factual}, we adhered to the following labelling
guidelines. In the cases where the claim was ``mostly true'', we decided to
label these as \texttt{factual}. When the claim was deemed ``half true'' or
``half false'' we labelled these as \texttt{misinformation}, with the
justification that a statement containing a significant part of false
information should be deemed as being misleading. When there was no clear
verdict then the verdict was labelled as \texttt{other}. This happens when the
answer is not known, or when the verdict is merely discussing the issue rather
than assessing the veracity of the claim. The claims with the \texttt{other}
label were not included in the final dataset.

To be able to properly deal with the multilingual verdicts, we attempted two
approaches: (1) Translate them into English and use a pre-trained English
language model; (2) Embed them using a pre-trained multilingual language model.

For the first approach, we used the Google Cloud Translation API\footnote{See
\url{https://cloud.google.com/translate/docs/reference/rest/}.} to translate
the verdicts into English and train a model to classify these translated
verdicts. The verdict classifier is based on the \texttt{roberta-base} model
\citep{liu2019roberta}, with an attached classification module. This
classification module consists of 10\% dropout, followed by a linear layer with
hidden size 768, a \texttt{tanh} activation, another 10\% dropout layer, and
finally a projection down to the three classes.

The model was trained on the dataset further augmented with casing.
Specifically, we converted all the verdicts in the training set to lower case,
upper case, title case and first letter capitalised. This resulted in a
training dataset of 10,000 verdicts. We manually labelled 1,000 further
verdicts to use as the test dataset. These verdicts were not deduplicated, to
ensure that their distribution matches the true one. The model was trained for
10 epochs using the \texttt{AdamW} optimizer \citep{loshchilov2017decoupled}
with a learning rate of 2e-5, and a batch size of 8.\footnote{This was done
using the \texttt{transformers} \citep{wolf2020transformers} and
\texttt{PyTorch} \citep{paszke2019pytorch} frameworks} The model achieved
a macro-F1 score of 0.99 among the \texttt{misinformation} and
\texttt{factual} classes, and 0.92 if the \texttt{other} class is included.

For the second multilingual approach, we augmented the original (multilingual)
verdicts, both by translating all of the 2,500 unique verdicts into 65
languages, using the Google Cloud Translation API\footnote{See
\url{https://cloud.google.com/translate/docs/reference/rest/}.} as well as
applying the casing augmentation as described above, as finetuning a
multilingual model directly on the original verdicts resulted in poor
performance for the minority languages. The resulting dataset consisted of
roughly 5 million verdicts, and we finetuned the \texttt{xlm-roberta-base}
model \cite{conneau2020unsupervised} for 4 epochs on the dataset with the same
hyperparameters as the model trained on the English-only dataset. On the same
test dataset of 1,000 multilingual verdicts, this multilingual model
achieved a macro-average F1-score of 0.98 among the \texttt{misinformation} and
\texttt{factual} classes, and 0.85 if the \texttt{other} class is included.

As the English-only model was marginally better than the multilingual model, we
opted to use that in building the dataset. However, we appreciate the
convenience of not having to translate the verdicts, so we release both the
English-only and multilingual verdict classifiers on the Hugging Face
Hub\footnote{See \url{https://hf.co/saattrupdan/verdict-classifier-en} and
\url{https://hf.co/saattrupdan/verdict-classifier}.}.

See Table~\ref{tab:verdict-clf} for some examples of the verdicts and resulting
predicted verdicts. With the performance satisfactory, we then used the model
to assign labels to all of the plaintext verdicts in the dataset.

\begin{table}[ht!]
	\caption{Sample predictions of the verdict classifier.}
    \begin{center}
        \scriptsize
        \begin{tabular}{c|c|c}
            \toprule
            \texttt{factual}&
            \texttt{misinformation}&
            \texttt{other}\\

            \midrule

            \wrapcenter{True} &
            \wrapcenter{False} &
            \wrapcenter{Satire} \\

            \wrapcenter{Correct Attribution} &
            \wrapcenter{Misleading} &
            \wrapcenter{Landmarks} \\

            \wrapcenter{Broadly correct.} &
            \wrapcenter{Mostly false} &
            \wrapcenter{Questionable} \\


            \wrapcenter{According to the most recent \\
                        data, this is about right} &
            \wrapcenter{Pants on fire} &
            \wrapcenter{More complex than that} \\

            \wrapcenter{This is correct for relative \\
                        poverty in the UK, measured \\
                        after housing costs in \\
                        2015/16. It’s a smaller \\
                        other measures of poverty.} &
            \wrapcenter{Three Pinocchios} &
            \wrapcenter{This video filmed in \\
                        Equatorial Guinea shows \\
                        a student attacking \\
                        one of his teachers} \\



            \bottomrule

        \end{tabular}
        \label{tab:verdict-clf}
    \end{center}
\end{table}

\subsection{Twitter}
\label{sec:twitter}
From the claims and verdicts, we next collected relevant social media data.
This data was collected from Twitter\footnote{\url{https://www.twitter.com}}
using their Academic Search
API\footnote{\url{https://developer.twitter.com/en/docs/twitter-api/tweets/search/api-reference/get-tweets-search-all}},
where we aimed to collect as many relevant Twitter threads that shared and
discussed content related the claims obtained through the method described in
Section~\ref{sec:claimcollection}.

\subsubsection{Data Collection}
\label{sec:twitterdatacollection}
From each claim, we first extracted the top 5 keyphrases\footnote{This was done
using the KeyBERT \citep{grootendorst2020keybert} package together with the
\texttt{paraphrase-multilingual-MiniLM-L12-v2} model from the Sentence
Transformer package \cite{reimers2019sentencebert}.}, with a keyphrase being
either one or two words from the claim, whose associated sentence embedding has
a large cosine similarity to the sentence embedding of the entire claim.

We then queried the Twitter Academic Search API for the first 100 results for
each of the five keyphrases, where we imposed that the tweets should not be
replies\footnote{Imposed with the query \texttt{-(is:reply OR is:retweet OR
is:quote)}.}, they had to share either a link or an image\footnote{Imposed
with the query \texttt{(has:media OR has:links OR has:images)}.} and they had
to have been posted at most three days prior to the claim date and at the
latest on the claim date itself. The idea behind this is to obtain as high a
recall as possible, i.e. obtaining as many of the potentially relevant tweets
as possible, from which we can filter the irrelevant tweets. From the Twitter
API we requested the tweets, users as well as media shared in the tweets. This
resulted in approximately 30 million tweets.


\subsubsection{Data Processing}
\label{sec:twitterdataprocessing}
As our goal with a automatic misinformation detection system is to be able to
act on stories shared on social media before they go viral, we decided to
filter the tweets to only keep the ones that have gained a reasonable amount of
interest. We measure such interest in terms of `retweets', and we chose a
minimum of 5 retweets to be a conservative threshold, which reduced the number
of tweets by 90\%, leaving about 2.5 million tweets.

From these we then extracted all the URLs and hashtags from the tweets, as well
as from the descriptions of each user. If the URL pointed to an article, we
downloaded the title, body and top image\footnote{This was done using the
\texttt{newspaper3k} Python library \citep{ouyang2013newspaper3k}.}. We also
extracted the hyperlinks of all images shared by the tweets. All of this
information was then populated in a graph database\footnote{We used the Neo4j
framework, see \url{https://neo4j.com/}.} with approximately 17 million nodes
and 50 million relations. Uniqueness constraints were imposed on all nodes.

\begin{figure*}
    \begin{center}
        \includegraphics[scale=0.32]{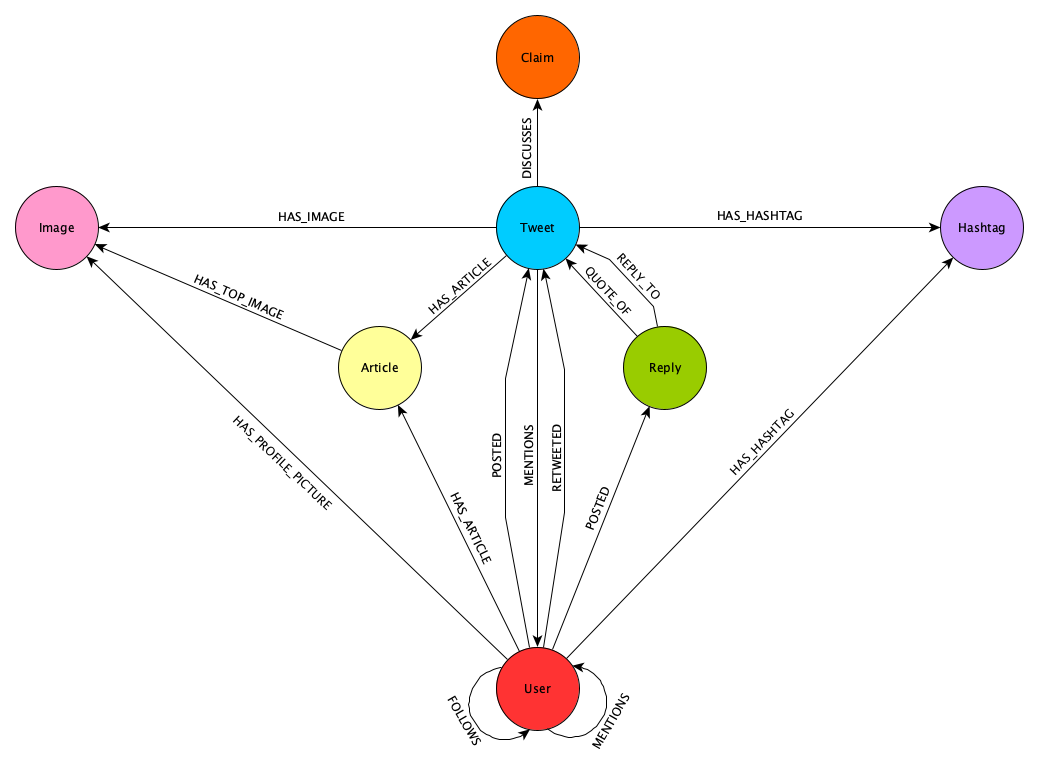}
    \end{center}
    \caption{The graph schema of the \datasetname dataset.}
    \label{fig:metagraph}
\end{figure*}

\subsection{Data Linking}
\label{sec:datalinking}
From the database of tweets, the next task was to find all the Twitter threads
that were relevant to each claim.
As the tweets, claims and articles were multilingual, we again had to make the
same choice as in Section~\ref{sec:claimdataprocessing}: either embed the texts
with a multilingual language model, or translate them to English and use an
English language model. We did both, with the translation being vastly
superior, as the multilingual model seemed to ``collapse'' on texts in certain
languages like Thai, Tamil, Telugu, Bengali and Malayalam. Translating texts
always comes with a risk of losing context, but as our goal was to find tweets
that were discussing a claim at hand, we argue that a translated text will
still be able to carry that information. The translation was done with the
Google Translation API, as with the verdicts in
Section~\ref{sec:claimcollection}.

Prior to embedding the approximately one million articles we first summarised
the concatenated title and abstract, using the \texttt{BART-large-CNN}
transformer \citep{lewis2019bart}. This was done to enable embedding of
additional tokens from the article content, as the Sentence Transformers have a
limit of 512 tokens with the summarisation model being able to process 1,024
tokens. We then embedded these summarised articles along with the claims and
tweets, all using the \texttt{paraphrase-mpnet-base-v2} Sentence Transformer
\cite{reimers2019sentencebert}.

Computing cosine similarities between every tweet-claim and article-claim pair
would be computationally unfeasible. Instead, we grouped the claims in batches
of size 100, fetched all the tweets and articles that were posted from three
days prior to one of the claims in the batch up until the claim date, and
computed cosine similarities between these\footnote{This was done using the
\texttt{PyTorch} framework \citep{paszke2019pytorch}.}.

\ifpreprint The resulting cosine similarity distribution can be found in the
appendix.\fi We decided to release three datasets, corresponding to the three
thresholds 0.7, 0.75 and 0.8. These thresholds were chosen based on a
qualitative evaluation of a subset of the linked claims; see examples of such
linked claims at various thresholds in Table~\ref{tab:data-linking}. The lower
threshold dataset is of course larger, but also contains more label noise,
whereas the higher threshold dataset is considerably smaller, but with higher
quality labels. See various statistics of these datasets in
Table~\ref{tab:statistics}.

\begin{table*}[ht!]
	\caption{Examples of claim-article linking.}
    \begin{center}
        \begin{tabular}{cccc}
            \toprule
			Translated Claim & Translated Title & Article URL & Similarity \\

            \midrule



            \wrapcentertiny{Google removed the term \\
                            ``Palestine'' from Google Maps} &
            \wrapcentertiny{Google and Apple remove \\
                            Palestine from their maps} &
            \wrapcentertiny{\url{https://bit.ly/mumi-3}} &
            \wrapcentertiny{84.93\%} \\

            \wrapcentertiny{China loses control of part \\
                            of its space rocket, and it \\
                            will soon fall to Earth.} &
            \wrapcentertiny{Heads Up! A Used Chinese \\
                            Rocket Is Tumbling Back to \\
                            Earth This Weekend.} &
            \wrapcentertiny{\url{https://bit.ly/mumi-4}} &
            \wrapcentertiny{80.47\%} \\

            \wrapcentertiny{Photo shows Aung San Suu Kyi \\
                            being detained during a \\
                            military coup on February \\
                            1, 2021} &
            \wrapcentertiny{Myanmar's army detains Aung \\
                            San Suu Kyi and government \\
                            leaders in a possible coup} &
            \wrapcentertiny{\url{https://bit.ly/mumi-5}} &
            \wrapcentertiny{75.03\%} \\

            \wrapcentertiny{One of the nurses who made the \\
                            Pfizer-BioNtech vaccine immediately \\
                            fainted from a side effect of the \\
                            vaccine. Also, the nurse who \\
                            fainted after having just been \\
                            vaccinated is dead.} &
            \wrapcentertiny{Live Nurse Faints After Being \\
                            Vaccinated Against Covid-19!} &
            \wrapcentertiny{\url{https://bit.ly/mumi-6}} &
            \wrapcentertiny{70.29\%} \\

            \wrapcentertiny{Americans Need WHO COVID-19 \\
                            Vaccine Card for International \\
                            Travel} &
            \wrapcentertiny{‘Vaccine passport’ will define \\
                            tourism in the world, but \\
                            countries bar some immunizers} &
            \wrapcentertiny{\url{https://bit.ly/mumi-7}} &
            \wrapcentertiny{65.30\%} \\




            \bottomrule

        \end{tabular}
        \label{tab:data-linking}
    \end{center}
\end{table*}

\subsection{Data Enrichment}
\label{sec:dataenrichment}
From the resulting Twitter posts linked to the claims as described in
Section~\ref{sec:datalinking}, we next queried Twitter for the surrounding
context of these posts. We retrieved a sample of 100 users that retweeted the
tweets, 100 users who followed the authors of the tweets, 100 users who were
followed by the authors of the tweets, 500 users who posted a reply to the
tweets and all users who was mentioned in the tweets. For each of these users,
we queried Twitter for their recent 100 tweets.

\section{Dataset Description}
\label{sec:datasetdescription}
Given the scale and diversity of the data collected, it is not possible to
succinctly provide a thorough analysis, which we leave to future work, and
other researchers interested in exploring and using our dataset. Nonetheless,
we will provide a preliminary analysis of various aspects of the dataset.

As mentioned in Section~\ref{sec:datalinking}, we release three datasets,
corresponding to the cosine similarity thresholds 0.7, 0.75 and 0.8. The
statistics of the datasets can be found in Table~\ref{tab:statistics}. Note the
heavy class imbalance of the datasets, which is likely due to the fact that
fact-checking organisations are more interested in novel claims, and these tend
to favour misinformation \citep{vosoughi2018spread}. A common way to fix this
issue \citep{li2020mmcovid, cui2020coaid} is to collect news articles from
``trusted sources'' and use tweets connected to these as a means to increase
the \texttt{factual} class. However, as these will likely arise from a
different distribution than the rest of the datasets (they might not be novel
claims, say), we decided against that and left the dataset as-is. We have
instead released the source code we used to collect the dataset, \trawlname,
which can be used to collect extra data, if needed\footnote{This can be found
at \dataseturl.}.

To adhere to the terms and conditions of Twitter, the dataset will only contain
the tweet IDs and user IDs, from which the tweets and the user data can be
collected via the Twitter API using our \pkgname package (see
Section~\ref{sec:muminpackage}). Further, to comply with copyright restrictions
of the fact-checking websites, we do not release the claims themselves.
Instead, we release keyphrases, obtained as described in
Section~\ref{sec:clusteranalysis}. The datasets thus contain the tweet IDs,
user IDs and claim keywords, as well as the \texttt{POSTED}, \texttt{MENTIONS},
\texttt{FOLLOWS}, \texttt{DISCUSSES} and \texttt{IS\_REPLY\_TO} relations,
shown in Figure~\ref{fig:metagraph}. From these, the remaining part of the
dataset can be built by using our \pkgname package, see
Section~\ref{sec:muminpackage}.

\subsection{Claim Topic Clusters}
\label{sec:clusteranalysis}
\begin{figure}[h!]
    \centering
     \includegraphics[scale=.3]{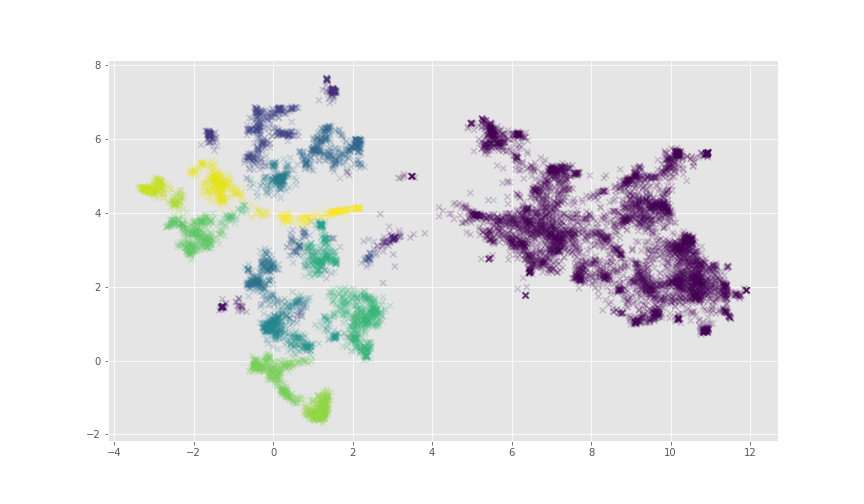}
    \caption{UMAP projection of the claim text embeddings. The large cluster on
             the right corresponds to COVID-19 related claims.}
    \label{fig:clusters}
\end{figure}

We performed clustering on embeddings of the claim text in order to extract
higher level topics or events from the claims. Using a \texttt{UMAP}
\citep{mcinnes2018umap} projection of embeddings of the claims and
\texttt{HDBSCAN} \citep{mcinnes2017accelerated}, a hierarchical density based
clustering algorithm, we were able to discover 26 clusters based on the claim
text. We optimized the hyperparameters of the projection as well as clustering
algorithms\footnote{This optimization resulted in the hyperparameters
\texttt{n\_neighbors}=50, \texttt{n\_components}=100,
\texttt{random\_state}=4242 and \texttt{metric}='cosine' for \texttt{UMAP},
and \texttt{min\_samples}=15 and \texttt{min\_cluster\_size}=40 for
\texttt{HDBSCAN}. This was done using the Python packages
\texttt{scikit-learn} \cite{pedregosa2011scikit} and \texttt{hdbscan}
\cite{mcinnes2017hdbscan}.}, achieving a silhouette coefficient of 0.28. The
clusters can be seen in Figure~\ref{fig:clusters}.

To provide context for each cluster, we concatenated the claims in each cluster
and extracted keyphrases from each cluster\footnote{This was done using the
KeyBERT library \citep{grootendorst2020keybert} on embeddings produced by a
Sentence Transformer \texttt{paraphrase-multilingual-MiniLM-L12-v2}
\cite{reimers2019sentencebert}.}. From these, it is apparent that the claims
can be clustered into diverse topics, ranging from COVID-19 (a cluster of
approximately half of all claims), to topics ranging from natural disasters to
national and international political and social events.

\section{The \pkgname Package}
\label{sec:muminpackage}

As we can only release the tweet IDs and user IDs to adhere to Twitter's terms
of use, we have built a Python package, \pkgname, to enable compilation of the
dataset as easily as possible. The package can be installed from PyPI using the
command \texttt{pip install mumin}, and the dataset can be compiled as follows:

\begin{verbatim}
>>> from mumin import MuminDataset
>>> dataset = MuminDataset(bearer_token, size='small')
>>> dataset.compile()
\end{verbatim}

Here \texttt{bearer\_token} is the Twitter API bearer token, which can be
obtained from the Twitter API website. The \texttt{size} argument determines
the size of the dataset to load and can be set to `small', `medium' or `large'.
Further, there are many arguments included in the \texttt{MuminDataset}
constructor which controls what data to include in the dataset. For instance,
one can set \texttt{include\_tweet\_images} to \texttt{False} to not include
any images\footnote{See \url{https://mumin-build.readthedocs.io} for a full
list of arguments.}.

With the dataset compiled, the graph nodes can be accessed through
\texttt{dataset.nodes} and the relations can be accessed through
\texttt{dataset.rels}. A convenience method \texttt{dataset.to\_dgl} returns a
heterogeneous graph object to be used with the DGL library \cite{wang2019dgl}.

We have built a tutorial on how to use the compiled dataset, including building
different classifiers. We also release the source code for the \pkgname
package\footnote{The tutorial and all the source code can be accessed through
\dataseturl.}.

\section{Model Performance}
\label{sec:modelperformance}

\subsection{Dataset Splits}
\label{sec:datasetsplits}
To enable consistent benchmarking on the dataset, we provide train-val-test
splits of the data. These have been created such that the splits are covering
distinct events, identified by the claim clusters in
Section~\ref{sec:clusteranalysis}. This is done as follows. We start by sorting
all the claim clusters by size, in ascending fashion. We next add clusters into
the validation set until at least 10\% of the claims have been included. We
then add clusters into the test set until at least 10\% of the claims have been
included, and the remaining clusters constitutes the training set. Statistics
for each of the splits can be found in Table~\ref{tab:splitstatistics}, which
shows that we still roughly maintain the label balance throughout all the
dataset splits.

\begin{table*}[ht!]
	\caption{Dataset split statistics}
    \begin{center}
        \begin{tabular}{c|ccc|ccc|ccc}
            \toprule
			Dataset & \%Train & \%Val & \%Test & \%MisinfoTrain & \%MisinfoVal &
                \%MisinfoTest & \#ClustersTrain & \#ClustersVal &
                \#ClustersTest \\
			\midrule
			\datasetname-large & 78.52\% & 11.39\% & 10.09\% & 94.37\% & 96.73\% &
                95.92\% & 8 & 21 & 8 \\
			\datasetname-medium & 76.98\% & 11.61\% & 11.41\% & 93.79\% & 96.73\% &
                94.46\% & 7 & 18 & 7 \\
			\datasetname-small & 77.90\% & 11.35\% & 10.75\% & 91.82\% & 97.15\% &
                94.42\% & 7 & 15 & 6 \\
            \bottomrule
        \end{tabular}
        \label{tab:splitstatistics}
    \end{center}
\end{table*}

\subsection{Baseline Models}
\label{sec:baselinemodels}
The \datasetname dataset lends itself to several different classification
tasks, relating the various modalities to the verdicts of the associated claims
(\texttt{misinformation} or \texttt{factual}). As mentioned in
Section~\ref{sec:introduction}, we have chosen to provide baselines related to
the following two tasks:

\begin{enumerate}
    \item Given a claim and its surrounding subgraph extracted from
        social media, predict whether the verdict of the claim is
        \texttt{misinformation} or \texttt{factual}.
        We name this task ``claim classification''.
    \item Given a source tweet (i.e., not a reply, quote tweet or retweet) to
        be fact-checked, predict whether the tweet discusses a claim whose
        verdict is \texttt{misinformation} or \texttt{factual}. We name this
        task ``tweet classification''.
\end{enumerate}

We implement several baseline models to demonstrate the predictive power of the
different modalities for these tasks. Firstly, we implement the \texttt{LaBSE}
transformer model from \cite{feng2020language} with a linear classification
head, and apply this model directly to the claims and the source tweets,
respectively. We also benchmark a version of this model where the transformer
layers are frozen, and name this model \texttt{LaBSE-frozen}. Secondly, we
implement the vision transformer (\texttt{ViT}) model from
\cite{dosovitskiy2020image}, also with a linear classification head, and apply
this to the subset of the tweets that include images (preserving the same
train/val/test splits).

As for a graph baseline, we implement a heterogeneous version of the
\texttt{GraphSAGE} model from \cite{hamilton2017inductive}, as follows. For
each node, we sample 100 edges of each edge type connected to it (in any
direction), process each of the sampled neighbouring nodes through a
\texttt{GraphSAGE} layer, and sum the resulting node representations. Finally,
layer normalisation \cite{ba2016layer} is applied to the aggregated node
representations. The baseline model contains two of these graph layers. This
graph baseline is trained on \datasetname\ without profile images, article
images and timelines (i.e., tweets that users in our graph have posted, which
are not directly connected to any claim)\footnote{Note that, as the graph
baseline has two layers, leaving these out does not change the claim
classification score, only potentially the tweet classification score.}.
We call this baseline model \texttt{HeteroGraphSAGE}.

See Table~\ref{tab:claimbaselines} and \ref{tab:tweetbaselines} for an overview
of the performance of each of these models. We see that both
tasks are really challenging, with the \texttt{HeteroGraphSAGE} model achieving
the best performance overall, but with the text-only \texttt{LaBSE} model not
far behind. We note that the \texttt{HeteroGraphSAGE} model only makes two
``hops'' through the graph, meaning that it is not able to capture all the
information that is present in the graph. Increasing the number of hops
resulted in poorer performance, which is the well-known ``oversmoothing''
problem \cite{li2018deeper,zhou2020graph}.

We have created an online leaderboard containing the results of these
baselines and invite researchers to submit their own models. We release all
the source code we used to conduct the baseline experiments.\footnote{See
\dataseturl\ for both the leaderboard and the baseline repository.}.

\begin{table}
    \caption{Baseline test performance on the claim classification task,
             measured in macro-average F1-score (larger is better). Best result
             for each dataset marked in bold.}
    \begin{center}
        \begin{tabular}{l|ccc}
            \toprule
            Model & \datasetname-small & \datasetname-medium &
                \datasetname-large \\
            \midrule
            Random &
                40.07\% &
                38.96\% &
                38.79\% \\
            Majority class &
                47.56\% &
                48.06\% &
                48.13\% \\
            \texttt{LaBSE-frozen} &
                57.50\% &
                54.10\% &
                55.00\% \\
            \texttt{LaBSE} &
                $\mathbf{62.55\%}$ &
                55.85\% &
                57.90\% \\
            \texttt{HeteroGraphSAGE} &
                57.95\% &
                $\mathbf{57.70\%}$ &
                $\mathbf{59.80\%}$ \\
            \bottomrule
        \end{tabular}
    \end{center}
    \label{tab:claimbaselines}
\end{table}

\begin{table}
    \caption{Baseline test performance on the tweet classification task,
             measured in macro-average F1-score (larger is better). Best result
             for each dataset marked in bold. Note that the \texttt{ViT} model
             is only trained and evaluated on the subset of the tweets
             containing images.}
    \begin{center}
        \begin{tabular}{l|ccc}
            \toprule
            Model & \datasetname-small & \datasetname-medium &
                \datasetname-large \\
            \midrule
            Random &
                37.18\% &
                37.72\% &
                36.90\% \\
            Majority class &
                48.77\% &
                48.56\% &
                48.87\% \\
            \texttt{ViT} &
                53.20\% &
                52.00\% &
                48.70\% \\
            \texttt{LaBSE} &
                54.50\% &
                $\mathbf{57.45\%}$ &
                52.80\% \\
            \texttt{HeteroGraphSAGE} &
                $\mathbf{56.05\%}$ &
                54.10\% &
                $\mathbf{61.45\%}$ \\
            \bottomrule
        \end{tabular}
    \end{center}
    \label{tab:tweetbaselines}
\end{table}

\section{Discussion}
\label{sec:discussion}

\subsection{Representative Data Splits}
\label{sec:datasetsplits}
In the field of automatic misinformation detection, the splitting of the
dataset into training/validation/test datasets is usually done uniformly at
random
\cite{wang2017liar,zubiaga2017exploiting,shu2020fakenewsnet,thorne2018fever,li2020mmcovid,dou2021upfd}.
However, we argue that the main purpose of such a system is to be able to
handle \textit{new} events in which misinformation occurs, and therefore our
dataset splits should reflect this. We conduct an experiment in which we
analyse the performance differences of the baseline models if we had split the
\datasetname dataset at random.

Concretely, we repeat two of our baselines on the random splits: the
\texttt{LaBSE} classifier and the \texttt{HeteroGraphSAGE} model. Call the
resulting models \texttt{LaBSE-random} and \texttt{HeteroGraphSAGE-random}.

On the tweet classification task, the \texttt{LaBSE-random} model achie-
ved a macro-average F1-score of 71.10\% and 73.6\% on
\datasetname-\texttt{small} and \datasetname-\texttt{medium}, respectively. The
\texttt{HeteroGraphSAGE}-\texttt{random} model achieved 74.90\%,
89.50\% and 79.80\% on \datasetname-\texttt{small},
\datasetname-\texttt{medium} and \datasetname-\texttt{large}, respectively. We
see that the scores are drastically higher for these ``random models'' on this
task compared to the results of the baseline models, as can be seen in
Table~\ref{tab:tweetbaselines}.

For the claim classification task, the \texttt{LaBSE-random} model achie- ved a
macro-average F1-score of 58.85\%, 62.80\% and 61.50\% on
\datasetname-\texttt{small}, \datasetname-\texttt{medium} and
\datasetname-\texttt{large}, respectively. On this task, the
\texttt{HeteroGraphSAGE}-\texttt{random} model achieved 48.50\%, 61.40\% and
62.55\%, respectively. There is not as big of a difference between these
``random models'' and our baselines as with the tweet classification task, as
can be seen from Table~\ref{tab:claimbaselines}, but the results are still
marginally better than the baseline models.

This shows that having realistic splits of the dataset is important to guide
our algorithm development in the right direction, especially in the field of
automated misinformation detection, where we are interested in generalisability
to new real-world events.

\subsection{Limitations}
\label{sec:limitations}
Due to the automated linking procedure between facts and tweets, and facts and
articles, erroneous labels exist. Nonetheless, this can be somewhat addressed
by selecting higher similarity thresholds, as can be seen in
Table~\ref{tab:data-linking}. We did not make any judgements as to the
impartiality or correctness of any of the verdicts provided by the
fact-checking organisations. Therefore, this dataset may contain verdicts
(labels) that are contentious or inaccurate. As a potential remedy, we provide
the fact-checking organisation responsible for each claim and verdict. As the
verdicts of each claim from fact-checking organisations are provided in
unstructured freetext, we resorted to a machine learning model to classify each
verdict into one of three categories, \texttt{factual}, \texttt{misinformation}
or \texttt{other}. While we obtained a high performance on a test set, it is
likely some verdicts may have been misclassified. As we do not distribute raw
social network data, but instead provide code to retrieve it, this means that
the dataset is truly dynamic such that if a user deletes a tweet or their
account, their data will not be retrievable from the Twitter API. This makes
reproducible research, if it involves the contents of the social network data,
challenging.

%

\subsection{Ethical Considerations}
\label{sec:ethicalconsiderations}
It is accepted that there are online harms associated with misinformation.
Unfortunately, when the number of posts made on social networks daily is
considered, the problem exists at a scale where manual curation is
exceptionally difficult, thus motivating the use of automated methods to assist
in the detection of misinformation online. These methods tend to utilise
machine learning, and therefore typically require the collection of large
amounts of data upon which to train the model. While the goal with such data
collection is to combat an online harm, there is, understandably, ethical
considerations related to the potential harms caused from the collection and
use of large online datasets of social data, text data, and media data. A major
factor for consideration is with respect to the collection of social network
data, and the fact that this data is generated from users of the social
network. The data collected in this dataset consists of only public Twitter
data, accessed through the official Twitter Academic API. While the users of
Twitter, in making their posts public, may expect their posts to be visible, in
accordance with the Twitter developer terms, we do not include the raw
collected data. Instead, we make available only tweet and user IDs, with
associated code to `hydrate’ them (i.e. retrieve the full tweet and user data).
Therefore, if a user deletes a tweet, or deletes their account, it will be no
longer possible to retrieve the deleted data from what we have released. Thus,
while we expect that data may disappear over time as a result, this trade-off
is required. The ethics of this work has been approved, both by the University
of Bristol Faculty of Engineering Research Ethics Committee (ref: 116665), as
well as by the Ethics Board at the National Research Centre on Privacy, Harm
Reduction and Adversarial Influence Online (REPHRAIN).

\section{Licenses}
\label{sec:licenses}
We release the three versions of the \datasetname under the Creative Commons
Attribution-NonCommercial 4.0 International license (CC BY-NC 4.0). The code,
which includes the \pkgname package, the data collection and linking system
\trawlname, as well as the repository containing the baselines, are all
released under the MIT license.\footnote{See \dataseturl\ for both \pkgname,
\trawlname\ and the baselines.}

\section{Conclusion}
In this paper we presented  \datasetname , which consists of a large scale
graph misinformation dataset that contains rich social media data (tweets,
replies, users, images, articles, hashtags) spanning 21 million tweets
belonging to 26 thousand Twitter threads, each of which have been semantically
linked to 13 thousand fact-checked claims across dozens of topics, events and
domains, in 41 different languages, spanning more than a decade. We also
presented a data collection and linking system, \trawlname. The freetext
multilingual verdicts were categorised into the consistent categories of
\texttt{factual} or \texttt{misinformation}, using a finetuned transformer
model which we also release. We further developed a Python package, \pkgname,
which enables simple compilation of the \datasetname as well as providing easy
export to Python graph machine learning libraries. Finally, we proposed and
provided baseline results for two node classification tasks; a) predicting the
veracity of a claim from its surrounding social context, and b) predicting the
likelihood that a tweet to be fact-checked discusses a misleading claim. The
baselines include text-only and image-only approaches, as well as a
heterogeneous graph neural network. We showed that the tasks are challenging,
with the highest macro-average F1-score being 62.55\% and 61.45\% for the two
tasks, respectively. The data, along with tutorials and a leaderboard, can be
found at \dataseturl.

\begin{acks}
    This research is supported by REPHRAIN: The National Research Centre on
    Privacy, Harm Reduction and Adversarial Influence Online, under UKRI grant:
    EP/V011189/1.
\end{acks}

\bibliographystyle{ACM-Reference-Format}
\bibliography{bibliography}
\clearpage

\ifpreprint

\appendix

\section{Supplementary Tables and Figures}

In Table~\ref{tab:bcp47} we show the 70 languages that were queried during our
data collection, and the resulting 41 languages, in bold, that made it to the
final dataset. In Table~\ref{tab:reviewers} we show the 115 fact checking
organisations whose fact checked claims have made it into the dataset. In
Table~\ref{tab:muminlargestats}, Table~\ref{tab:muminmediumstats} and
Table~\ref{tab:muminsmallstats} we show language statistics across the large,
medium and small versions of the dataset, respectively. In
Figure~\ref{fig:cosine-similarity} we show the cosine similarity distribution
of the tweet-claim pairs.

\setlength{\tabcolsep}{1pt}

\begin{table}[H]
    \caption{The 70 languages queried, with the 41 languages in \textbf{bold}
    present in the final dataset.}
    \begin{center}
        \begin{tabular}{llll}
            Amharic & Georgian & Lithuanian & Sinhala \\
            \textbf{Arabic} & \textbf{German} & \textbf{Macedonian} &
                \textbf{Slovak} \\
            Armenian & \textbf{Greek} & \textbf{Malayalam} & Slovenian \\
            Azerbaijani & \textbf{Gujarati} & \textbf{Malay} &
                \textbf{Spanish} \\
            Basque & Haitian Creole & \textbf{Marathi} & Swedish \\
            \textbf{Bengali} & \textbf{Hebrew} & Nepali & Tagalog \\
            Bosnian & \textbf{Hindi} & \textbf{Norwegian} & \textbf{Tamil} \\
            Bulgarian & \textbf{Hungarian} & \textbf{Oriya} & \textbf{Telugu} \\
            \textbf{Burmese} & Icelandic & \textbf{Panjabi} & \textbf{Thai} \\
            \textbf{Croatian} & \textbf{Indonesian} & Pashto &
                \textbf{Traditional Chinese} \\
            Catalan & \textbf{Italian} & \textbf{Persian} & \textbf{Turkish} \\
            \textbf{Czech} & \textbf{Japanese} & \textbf{Polish} & Ukranian \\
            \textbf{Danish} & Kannada & \textbf{Portuguese} & \textbf{Urdu} \\
            \textbf{Dutch} & \textbf{Kazakh} & Romanian & Uyghur \\
            \textbf{English} & Khmer & \textbf{Russian} & Vietnamese \\
            Estonian & \textbf{Korean} & \textbf{Serbian} & Welsh \\
            \textbf{Filipino} & Lao & \textbf{Simplified Chinese} \\
            Finnish & Latvian & Sindhi \\
            \textbf{French}
        \end{tabular}
        \label{tab:bcp47}
    \end{center}
\end{table}

\begin{table}[H]
    \caption{The distribution of the top languages in the \datasetname-large
    dataset.}
    \begin{center}
        \begin{tabular}{lrrr}
            \toprule
            \textbf{Language} &
            \textbf{Proportion} &
            \textbf{\#Claims} &
            \textbf{\%\texttt{misinfo}} \\

            \midrule

            English & 42.88\% & 5,538 & 92.85\% \\
            Portuguese & 10.98\% & 1,418 & 95.28\% \\
            Spanish & 8.26\% & 1,067 & 95.41\% \\
            Hindi & 6.16\% & 796 & 100.00\% \\
            Arabic & 4.34\% & 560 & 95.18\% \\
            French & 3.46\% & 447 & 97.99\% \\
            German & 2.91\% & 376 & 97.61\% \\
            Indonesian & 2.55\% & 329 & 99.70\% \\
            Italian & 2.33\% & 301 & 89.37\% \\
            Bengali & 2.26\% & 292 & 100.00\% \\
            Turkish & 2.19\% & 283 & 95.41\% \\
            Polish & 1.73\% & 224 & 83.48\% \\
            Other & 9.93\% & 1,283 & 95.49\% \\
            \bottomrule
        \end{tabular}
        \label{tab:muminlargestats}
    \end{center}
\end{table}

\begin{table}[H]
    \caption{The distribution of the top languages in the \datasetname-medium
    dataset.}
    \begin{center}
        \begin{tabular}{lrrr}
            \toprule
            \textbf{Language} &
            \textbf{Proportion} &
            \textbf{\#Claims} &
            \textbf{\%\texttt{misinfo}} \\

            \midrule

            English & 45.46\% & 2,530 & 92.29\% \\
            Portuguese & 10.75\% & 598 & 96.49\% \\
            Spanish & 7.82\% & 435 & 94.25\% \\
            Hindi & 6.50\% & 362 & 100.00\% \\
            Arabic & 4.40\% & 245 & 93.88\% \\
            French & 3.61\% & 201 & 97.51\% \\
            Italian & 3.04\% & 169 & 86.98\% \\
            German & 2.57\% & 143 & 97.90\% \\
            Indonesian & 2.07\% & 115 & 100.00\% \\
            Bengali & 1.99\% & 111 & 100.00\% \\
            Turkish & 1.90\% & 106 & 94.34\% \\
            Polish & 1.40\% & 106 & 80.77\% \\
            Other & 8.48\% & 472 & 97.03\% \\
            \bottomrule
        \end{tabular}
        \label{tab:muminmediumstats}
    \end{center}
\end{table}

\begin{table}[H]
    \caption{The distribution of the top languages in the \datasetname-small
    dataset.}
    \begin{center}
        \begin{tabular}{lrrr}
            \toprule
            \textbf{Language} &
            \textbf{Proportion} &
            \textbf{\#Claims} &
            \textbf{\%\texttt{misinfo}} \\

            \midrule

            English & 47.41\% & 1,035 & 90.34\% \\
            Portuguese & 10.86\% & 237 & 97.47\% \\
            Spanish & 7.42\% & 162 & 92.59\% \\
            Hindi & 6.92\% & 151 & 100.00\% \\
            Arabic & 4.90\% & 107 & 89.72\% \\
            Italian & 4.49\% & 98 & 86.73\% \\
            French & 3.71\% & 81 & 97.53\% \\
            Turkish & 1.83\% & 40 & 87.50\% \\
            German & 1.51\% & 33 & 100.00\% \\
            Indonesian & 1.51\% & 33 & 100.00\% \\
            Bengali & 1.42\% & 31 & 100.00\% \\
            Polish & 1.15\% & 25 & 80.00\% \\
            Other & 6.87\% & 150 & 96.00\% \\
            \bottomrule
        \end{tabular}
        \label{tab:muminsmallstats}
    \end{center}
\end{table}

\begin{table*}[h]
        \caption{The 115 fact-checking organisations present in the dataset.
        The numbers in parentheses indicate how many claims were processed from
        the website in total.}
    \begin{center}
        \footnotesize
        \begin{tabular}{lr | lr | lr}
            \toprule
            \textbf{Website} & \textbf{Claims included} &
            \textbf{Website} & \textbf{Claims included} &
            \textbf{Website} & \textbf{Claims included} \\

            \midrule

            politifact.com & 716 (7,865) & factcheck.kz & 90 (776) & thip.media & 19 (134) \\
            factcheck.afp.com & 581 (4,874) & correctiv.org & 87 (1,313) & scroll.in & 18 (73) \\
            boomlive.in & 407 (3,149) & faktograf.hr & 86 (680) & faktisk.no & 17 (640) \\
            factual.afp.com & 363 (2,913) & newschecker.in & 83 (1,143) & ici.radio-canada.ca & 17 (102) \\
            snopes.com & 361 (4,025) & fatabyyano.net & 77 (1,218) & fakenews.pl & 17 (163) \\
            misbar.com & 328 (4,641) & animalpolitico.com & 66 (850) & thejournal.ie & 16 (83) \\
            factly.in & 317 (4,113) & factcheck.thedispatch.com & 64 (177) & malayalam.factcrescendo.com & 15 (245) \\
            dpa-factchecking.com & 298 (1,474) & lemonde.fr & 62 (564) & factnameh.com & 15 (387) \\
            vishvasnews.com & 298 (5,974) & bol.uol.com.br & 62 (407) & factrakers.org & 13 (147) \\
            factcheck.org & 268 (2,312) & factcheckthailand.afp.com & 58 (252) & factograph.info & 12 (253) \\
            factuel.afp.com & 243 (2,710) & projetocomprova.com.br & 57 (406) & watson.ch & 11 (39) \\
            facta.news & 230 (1,196) & noticias.uol.com.br & 56 (693) & poynter.org & 9 (49) \\
            fullfact.org & 226 (3,302) & sprawdzam.afp.com & 54 (299) & br.de & 9 (121) \\
            thequint.com & 223 (1,084) & dogrulukpayi.com & 53 (641) & mygopen.com & 8 (440) \\
            observador.pt & 207 (1,284) & aap.com.au & 52 (365) & factcheckni.org & 8 (141) \\
            aajtak.in & 189 (1,539) & newsweek.com & 48 (196) & hindi.asianetnews.com & 8 (165) \\
            piaui.folha.uol.com.br & 187 (6,060) & tamil.factcrescendo.com & 47 (1,523) & abc.net.au & 7 (112) \\
            newtral.es & 178 (2,353) & periksafakta.afp.com & 47 (415) & liberation.fr & 7 (97) \\
            checamos.afp.com & 165 (1,073) & chequeado.com & 46 (1,689) & theconversation.com & 6 (54) \\
            polygraph.info & 157 (1,128) & nytimes.com & 44 (497) & telugu.newsmeter.in & 6 (280) \\
            aosfatos.org & 155 (1,795) & poligrafo.sapo.pt & 42 (3,496) & factchecker.in & 6 (32) \\
            teyit.org & 154 (2,421) & boombd.com & 39 (381) & open.online & 5 (23) \\
            usatoday.com & 154 (884) & fakty.afp.com & 38 (220) & bbc.co.uk & 5 (43) \\
            politica.estadao.com.br & 151 (1,632) & dailyo.in & 36 (729) & tenykerdes.afp.com & 5 (36) \\
            factcrescendo.com & 145 (896) & presseportal.de & 35 (466) & namibiafactcheck.org.na & 4 (36) \\
            thelogicalindian.com & 139 (994) & youturn.in & 35 (1,591) & factcheckmyanmar.afp.com & 4 (79) \\
            washingtonpost.com & 138 (1,304) & 20minutes.fr & 33 (255) & observers.france24.com & 4 (54) \\
            cekfakta.com & 135 (4,104) & altnews.in & 31 (4,996) & oglobo.globo.com & 4 (50) \\
            bangla.boomlive.in & 131 (1,640) & cbsnews.com & 30 (231) & buzzfeed.com & 2 (25) \\
            ellinikahoaxes.gr & 131 (1,120) & napravoumiru.afp.com & 29 (172) & bangla.aajtak.in & 2 (129) \\
            newsmeter.in & 127 (1,430) & semakanfakta.afp.com & 29 (198) & istinomer.rs & 2 (887) \\
            boatos.org & 125 (1,893) & faktencheck.afp.com & 27 (335) & verify-sy.com & 2 (56) \\
            maldita.es & 123 (1,063) & tjekdet.dk & 27 (481) & thewhistle.globes.co.il & 2 (65) \\
            colombiacheck.com & 118 (802) & cinjenice.afp.com & 26 (227) & azattyq.org & 1 (9) \\
            demagog.org.pl & 115 (3,181) & vistinomer.mk & 25 (370) & radiofarda.com & 1 (33) \\
            indiatoday.in & 115 (1,433) & tfc-taiwan.org.tw & 25 (1,077) & assamese.factcrescendo.com & 1 (40) \\
            healthfeedback.org & 111 (328) & factcheckkorea.afp.com & 24 (194) & tamil.newschecker.in & 1 (26) \\
            hindi.boomlive.in & 109 (1,372) & malumatfurus.org & 24 (731) \\
            cekfakta.tempo.co & 95 (1,142) & rappler.com & 24 (350) \\

            \bottomrule

        \end{tabular}
        \label{tab:reviewers}
    \end{center}
\end{table*}

\begin{figure}[H]
    \begin{center}
        \includegraphics[width=0.5\textwidth]{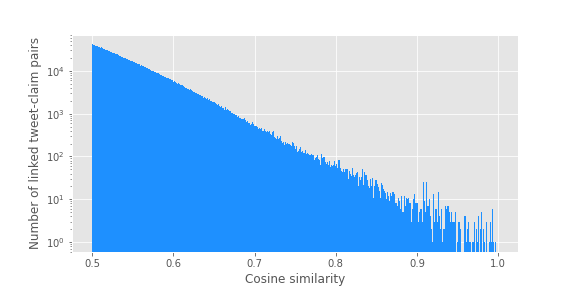}
    \end{center}
    \caption{The distribution of cosine similarities among tweet-claim pairs.}
    \label{fig:cosine-similarity}
\end{figure}

\fi

\end{document}